\documentclass{article}
\usepackage{amssymb}

\usepackage[final]{corl_2025} 
\usepackage{times}

\usepackage[numbers]{natbib}
\usepackage{multicol}

\usepackage{subfigure}
\usepackage{xcolor}
\usepackage{times}
\usepackage{epsfig}
\usepackage{amsmath}
\usepackage{amssymb}
\usepackage{mathrsfs}
\usepackage[ruled,vlined,linesnumbered]{algorithm2e}
\usepackage{wrapfig}
\usepackage{microtype}
\usepackage{multirow}
\usepackage{lineno}

\newcommand{\GGG}{\mathcal{G}}

\newcommand{\HHH}{\mathcal{H}}

\newcommand{\VVV}{\mathcal{V}}

\newcommand{\LLL}{\mathcal{L}}

\newcommand{\NNN}{\mathcal{N}}

\newcommand{\hh}{\mathbf{h}}

\newcommand{\state}{\mathbf{s}}

\newcommand{\vv}{\mathbf{v}}

\newcommand{\pos}{\mathbf{p}}
\newcommand{\goal}{\mathbf{g}}
\newcommand{\vel}{\mathbf{q}}

\newcommand{\aaa}{\mathbf{a}}

\newcommand{\ADJ}{\mathbf{E}}

\newcommand{\mm}{\mathbf{m}}

\newcommand{\WW}{\mathbf{W}}

\newcommand{\zz}{\mathbf{z}}
\newcommand{\ff}{\mathbf{f}}
\newcommand{\YY}{\mathbf{Y}}

\makeatother

\title{Subteaming and Adaptive Formation Control for Coordinated Multi-Robot Navigation}

\author{
    \textbf{Zihao Deng}$^{1*}$, 
    \textbf{Peng Gao}$^{2*}$, 
    \textbf{Williard Joshua Jose}$^1$, 
    \textbf{Maggie Wigness}$^{3}$, \\
    \textbf{John Rogers}$^{3}$\textbf{,} 
    \textbf{Brian Reily}$^{3}$\textbf{,} 
    \textbf{Christopher Reardon}$^4$\textbf{,} 
    \textbf{and Hao Zhang}$^1$\vspace{3pt}\\
    $^1$ University of Massachusetts Amherst \quad 
    $^2$ North Carolina State University \\
    $^3$ U.S. Army DEVCOM Army Research Laboratory \quad 
    $^4$ University of Denver\vspace{3pt}\\
    $^*$ Authors contributed equally to this paper\\
}

\begin{document}
\maketitle

\begin{abstract}
Coordinated multi-robot navigation is essential for robots to operate as a team in diverse environments. 
During navigation, robot teams usually need to maintain specific formations, such as circular formations to protect human teammates at the center. 
However, in complex scenarios such as narrow corridors, rigidly preserving predefined formations can become infeasible. 
Therefore, robot teams must be capable of dynamically splitting into smaller subteams and adaptively controlling the subteams to navigate through such scenarios while preserving formations.
To enable this capability, we introduce a novel method for \textit{SubTeaming and Adaptive Formation} (STAF), which is built upon a unified hierarchical learning framework:
(1) high-level deep graph cut for team splitting, (2) intermediate-level graph learning for facilitating coordinated navigation among subteams, 
and (3) low-level policy learning for controlling individual mobile robots to reach their goal positions while avoiding collisions.
To evaluate STAF, we conducted extensive experiments in both indoor and outdoor environments using robotics simulations and physical robot teams.
Experimental results show that STAF enables the novel capability for subteaming and adaptive formation control, and achieves promising performance in coordinated multi-robot navigation through challenging scenarios.
More details are available on the project website: \href{https://hcrlab.gitlab.io/project/STAF/}{https://hcrlab.gitlab.io/project/STAF}.
\end{abstract}



\section{Introduction}
\begin{wrapfigure}{R}{0.5\textwidth}
\vspace{-12pt}
\centering
\includegraphics[width=0.495\textwidth]{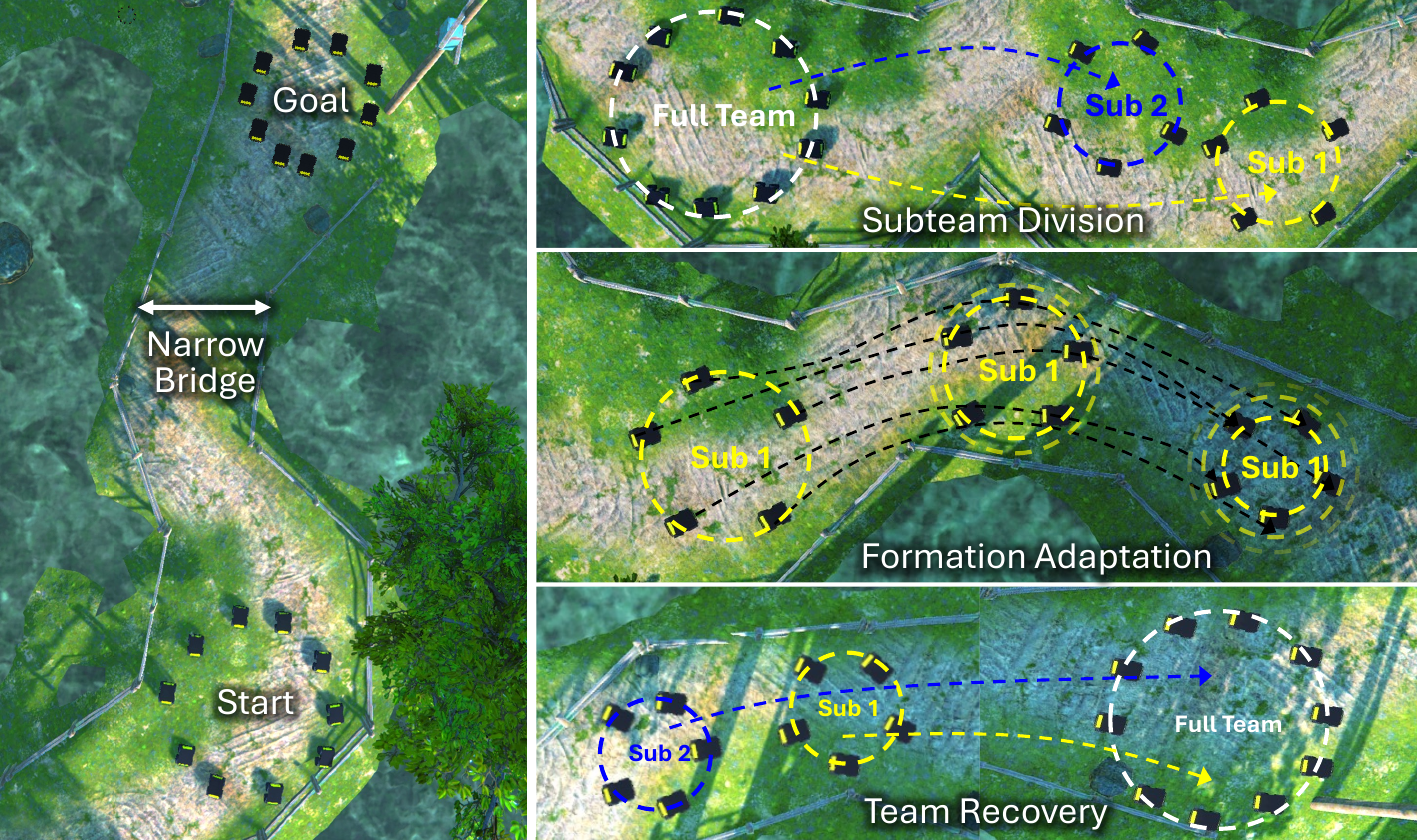}
\vspace{-15pt}
\caption{When a robot team  in circular formation encounters a bridge that is too narrow for the entire team to cross at once. The robots must divide into subteams, adapt their formations to navigate the bridge, and recovery the full team after crossing.}
\label{fig:motivation}
\vspace{-15pt}
\end{wrapfigure}
Multi-robot systems have attracted growing attention due to their
advantages, such as redundancy \cite{gao2023collaborative}, parallelism \cite{pinciroli2012argos}, and scalability \cite{balch2000social}. 
Coordinated multi-robot navigation is a fundamental capability that allows teams of robots to traverse environments in a synchronized manner and reach goal positions collectively \cite{singh2016navigation}.
This capability is crucial in real-world applications, such as search and rescue \cite{queralta2020collaborative, yang2020needs, baxter2007multi}, space exploration \cite{han2020cooperative, indelman2018cooperative}, and transportation \cite{amanatiadis2015avert, koung2021cooperative}.

During coordinated navigation, robots are often required to maintain mission-specific formation, such as circular formations for protection or line formations for coverage.
However, rigid adherence to predefined formations can hinder effective navigation in certain scenarios.
For instance, Figure \ref{fig:motivation} depicts a team of ten robots in a circular formation encountering a corridor too narrow for the entire team to pass through. 
Thus, the team must be capable of dynamically dividing into smaller units that operate both independently and cohesively (i.e., \textit{subteaming}) 
and controlling the subteams to pass through the narrow corridor while adaptively maintaining a specific formation (i.e., \textit{adaptive formation control}).

The importance of coordinated multi-robot navigation has driven the development of various techniques. 
Traditional approaches, including classical planning methods \cite{kuffner2000rrt}, game-theoretic methods \cite{tang2020multi, 9143181}, and optimization-based methods \cite{reily2020leading}, often face high computational costs. 
Recently, learning-based methods like deep neural networks \cite{goarin2024graph, zhang2023neural} and multi-agent reinforcement learning \cite{blumenkamp2022framework, hu2023graph} have been used for modeling, coordination, and navigation.
However, these methods have not addressed adaptive formation control, which is critical for narrow corridor traversal.
Subteaming methods, such as graph cuts for team division \cite{zhu2019distributed, reily2020representing} and mixed-integer programming for task allocation \cite{gao2023collaborative, jose2024learning, cardona2023temporal}, focus on team division alone and lack control over subteams or individual robots, which limits their effectiveness for coordinated navigation.

To address the challenges above and enable effective coordinated multi-robot navigation in complex scenarios where the entire robot team cannot pass through, we introduce a novel approach called \textit{SubTeaming and Adaptive Formation} (\textbf{STAF}), which offers new capabilities for subteam division, formation adaptation, and team recovery.
Specifically, we design a graph representation to encode a team of robots, 
where each node represents a robot along with its associated attributes, such as its position, velocity, goal, and distance to obstacles, and each edge represents the spatial relationships between pairs of robots.
Our STAF approach integrates three levels of robot learning into a hierarchical framework. 
At the high level, given the graph representation of a robot team, STAF performs deep graph cuts to divide the entire robot team into subteams. 
The intermediate level of STAF focuses on learning the coordination of these robot subteams for navigation, 
which develops a graph neural network with learnable message sharing to coordinate robots within a subteam, while generating graph embeddings to encode the subteam context. 
Finally, at the low level, given these embeddings, STAF employs reinforcement learning to learn a navigation policy that controls each individual robot to adaptively maintain subteam formation, reach the goal position, and avoid collisions.

Our primary contribution is the introduction of the novel STAF method to enable a new multi-robot navigation capability of subteaming and formation adaptation. 
The specific novelties include:
\begin{itemize}

        \item This work introduces one of the first problem formulations and learning-based  solutions for subteaming and formation adaptation in multi-robot coordinated navigation.
        It enables new multi-robot capabilities, including subteam division, formation adaptation, and team recovery,
        allowing a team of robots to navigate complex environments in a coordinated manner, particularly narrow corridors where maintaining original formation is infeasible.

        \item We introduce a novel hierarchical robot learning method that simultaneously integrates high-level deep graph cut for subteaming, intermediate-level graph learning for subteam coordination and adaptive formation control, and low-level individual robot control for collision-free navigation in complex environments. 

\end{itemize}




\section{Related Work}
\textbf{Hierarchical Learning for Robotics}
Hierarchical learning has shown promise in complex multi-robot tasks by providing a structured problem formulation that better aligns with multi-objective goals.
It also enhances modularity in model design, which improves interpretability and enables clearer evaluation of each level.
Applications include task allocation \cite{wang2021bi}, maintaining communication \cite{bettini2023heterogeneous}, path planning \cite{zhang2023bi, wang2024multi}, and consensus reaching \cite{feng2024hierarchical}.
Typically, the lower level handles individual control tasks such as obstacle avoidance \cite{bischoff2013hierarchical, jin2021hierarchical}.
The upper level focuses on team planning and coordination \cite{wohlke2021hierarchies, hu2020voronoi, zhu2022hierarchical,blumenkamp2022framework}.
However, applying hierarchical learning to formation adaptation and subteaming remains challenging due to the need for scalable team representations, dynamic adaptation, and efficient integration of formation control with flexible team reconfiguration.

\textbf{Coordinated Multi-Robot Navigation}
Learning-free methods rely on predefined formation strategies, such as leader-follower \cite{reily2020leading, singh2016navigation, wu2022leader, xiao2019leader} and virtual region methods \cite{8594438, abujabal2023comprehensive, roy2019virtual, alonso2019distributed}.
However, these rigid formations lack adaptability to environmental changes.
Learning-based methods, such as reinforcement learning (RL) \cite{han2022reinforcement, hacene2021behavior, blumenkamp2022framework, hu2023graph, 9197209, hacene2021behavior}, address this limitation by optimizing actions through environmental feedback.
Graph neural networks (GNNs) enhance team coordination and communication \cite{li2020graph, zhang2023neural}, enabling decentralized decision-making \cite{goarin2024graph, gao2024co}.
These approaches have been applied in areas such as connected autonomous driving \cite{gao2024collaborative, han2020cooperative}, area coverage \cite{ozkahraman2022collaborative}, and search-and-rescue missions \cite{queralta2020collaborative}.
However, none of these methods effectively address subteaming and formation adaptation in coordinated navigation, particularly in complex narrow corridors. 

\textbf{Subteaming in Multi-Robot Navigation and Task Allocation}
Subteaming increases the complexity of coordinated multi-robot navigation as it involves splitting, merging, and reformation based on tasks or environments.  
Graph-based methods \cite{zhu2019distributed,reily2020representing, luo2020behavior, gao2023collaborative} use graph partitioning and cutting to determine team division and merging, but often rely on explicit connectivity constraints.
Leader-follower methods \cite{roy2021exploration, swaminathan2015planning, reily2020leading} apply predefined hierarchy-based strategies but lack flexibility in dynamic environments.
Optimization-based approaches \cite{novoth2020distributed, jose2024learning, cardona2023temporal, calvo2024optimal} compute optimal assignments via mixed-integer programming. 
Heuristic-based methods \cite{guo2021spatial, guo2023efficient} use problem-specific heuristics to determine team formation and coordination strategies.
However, these methods focus on team division alone and lack control over subteams or individual robots. See Appendix for details \footnote{The appendix is available at our project website.}.

\section{Approach}\label{sec:approach}




\begin{figure*}[t]
\centering
\includegraphics[width=0.95\textwidth]{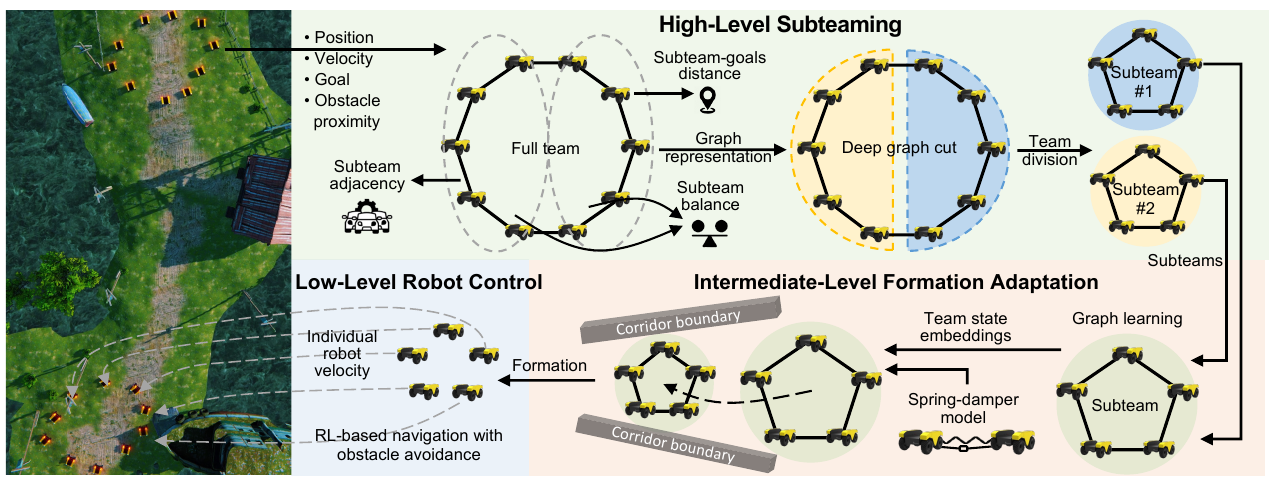}
\vspace{-3pt}
\caption{Overview of STAF, which integrates three levels of robot learning within a unified hierarchical learning framework to enable coordinated multi-robot navigation.
}
\label{fig:Motivation}
\vspace{-8pt}
\end{figure*}
\textbf{Problem Definition}
We discuss our STAF method that enables new multi-robot capabilities of subteaming and formation adaptation for coordinated multi-robot navigation.
An overview of STAF is illustrated in Figure \ref{fig:Motivation}. 
We represent a team of $n$ robots using an undirected graph $\GGG=\{\VVV,\ADJ\}$. 
In the node set $\VVV = \{\vv_1, \vv_2, \dots, \vv_n\}$,
each node $\vv_i=\{\pos_i, \goal_i, \vel_i\}$ consists of the attributes of the $i$-th robot, 
where $\pos_i=[p^{x}_i, p^{y}_i]$ denotes its position, 
$\goal_i=[g^x_i, g^y_i]$ denotes its goal position, 
and $\vel_i=[q^{x}_i, q^{y}_i]$ denotes its velocities along x and y directions.
The edge matrix $\ADJ = \{a_{i,j}\}^{n \times n}$ represents the spatial adjacency of the robots, 
where $a_{i,j}=1$, if the $i$-th robot and the $j$-th robot are within a radius; otherwise $a_{i,j}=0$.
We further define the state of the $i$-th robot $\state_i=[\pos_i, \goal_i, \vel_i, c_i]$ as the concatenation of the robot's attributes  
and the distance  $c_i$ between the robot and its closest obstacle. 
We define the action of the $i$-th robot as  $\aaa_i = [v^x_i, v^y_i]$,
where $v^x_i$ and $v^y_i$ denote the robot's velocities in the x and y directions, respectively. 

Our objective is to address the problems of subteaming and formation adaptation in the context of coordinated multi-robot navigation:
\begin{itemize}
    \item \textbf{Formation Adaptation:} The capability of a robot team or subteam to maintain a desired formation while dynamically adjusting their relative positions to safely and efficiently navigate through the unstructured environment toward their goal positions, particularly in challenging scenarios such as narrow corridors.

    
    \item \textbf{Subteaming}: The capability of a robot team with a specific formation to autonomously divide into subteams with the same formation type when navigating environments too narrow for the entire robot team. After successfully passing through, the subteams must merge back into the full team, restoring the original formation.

    
\end{itemize}


\textbf{High-Level Deep Graph Cut for Subteaming}
Given the graph  $\GGG$  as the representation of the robot team, we introduce a new deep graph cut approach at the high level of STAF to enable subteaming. 
We compute the embedding of the robot graph as $\HHH = \{\hh_i\} = \omega(\GGG)$, 
where $\hh_i$ is the embedding of the $i$-th robot and $\omega$ is a graph attention network \cite{velivckovic2017graph}. 
We project each node into a representation space by calculating $\mm_i = \WW^v \pos_i$,
where $\mm_i$ denotes the projected feature vector of the $i$-th node, and $\WW^v$ denotes the weight matrix.
Then, we compute the attention $\alpha_{i,j}$  from the $j$-th node to the $i$-th node as $\alpha_{i,j} =  \frac{ \exp\left(\text{ReLu} ([\WW^a\mm_i||\WW^a\mm_j])\right)}{\sum_{k\in \NNN(i)}\exp(\text{ReLu} ([\WW^a\mm_i||\WW^a \mm_k]))}$,
where $\text{ReLu}$ denotes the rectified linear unit activation function,  $\NNN(i)$ represents the set of adjacent nodes of the $i$-th node, $||$ denotes the concatenation operation, and $\WW^a$ represents the weight matrix. 
The attention $\alpha_{i,j}$ is obtained by computing the similarity of the $i$-th node with
its $j$-th adjacent nodes, followed by the SoftMax normalization. 
Then, the final embedding $\hh_i$ for the $i$-th node is computed through aggregating the embeddings of all its adjacent nodes as $\hh_i = \WW^h \mm_i + \sum_{j \in \NNN(i)} \alpha_{i,j} \WW^h \mm_j $,
where 
$\WW^h$ is the weight matrix.
We further utilize a multi-head mechanism \cite{velivckovic2017graph} after the attention layers to enable the network to capture a richer embedding representation.

Given $\HHH = \{\hh_i\} $, we formulate subteaming as a graph cut problem,
which partitions  the entire graph (representing the full team) into $m$ subgraphs (representing subteams).
In order to compute team division, we develop a classifier network $\tau(\HHH)$ consisting of  two fully connected linear layers followed by a SoftMax function,
which outputs the team division results as $\YY = \tau(\HHH) = \{y_{i,j}\}^{n \times m}$,
where $y_{i,j}$ is the probability of the $i$-th robot belonging to the $j$-th subteam,
and $m<n$.


To ensure that robots within the same subteam group together,
i.e., each robot is adjacent to its teammates within the same subteam,
we define a loss function that maximizes the adjacency of robots within each subteam 
as $\YY(1-\YY)^\top \ADJ$,
where $\YY(1-\YY)^\top$ calculates the probability that a pair of robots belong to different subteams,
and $\ADJ$ encodes the adjacency of the robots. 
In addition, we aim to maintain balance in the sizes of robot subteams, encouraging each subteam to have the same or a similar number of robots.
It can be mathematically modeled by a loss function $\sum_{j=1}^{m}\left(\sum_{i=1}^{n}y_{i,j}-\frac{n}{m}\right)$.
The term $\frac{n}{m}$ calculates the optimal size of balanced subteams (e.g., when $n=10$ and $m=2$, 
each subteam would consist of 5 robots). 
Furthermore, we model the mission objective of reaching the goal position by minimizing the overall distance between the subteams and their respective goal positions.
It can be mathematically defined as 
$\sum_{j=1}^{m} \big\| \frac{\sum_{i=1}^{n} y_{i,j}  \pos_i}{\sum_{i=1}^{n}y_{i,j}} - \frac{\sum_{i=1}^{n} y_{i,j}  \goal_i}{\sum_{i=1}^{n}y_{i,j}}\big\|_2$,
where $\frac{\sum_{i=1}^{n} y_{i,j}  \pos_i}{\sum_{i=1}^{n}y_{i,j}}$ denotes the center position of the $j$-th subteam and 
$\frac{\sum_{i=1}^{n} y_{i,j}  \goal_i}{\sum_{i=1}^{n}y_{i,j}}$ denotes the center position of the goal for the subteam. 

The high-level component of STAF performs an unsupervised graph cut to enable team division for subteaming by minimizing the following objective function:
\begin{equation}\label{eq:loss}
    \LLL_{st} = \overbrace{\YY (1-\YY)^\top \ADJ}^{\text{Subteam adjacency}} 
    + 
    \overbrace{\sum_{j=1}^{m} \left(\sum_{i=1}^{n}y_{i,j}-\frac{n}{m}\right)}^{\text{Subteam balance}}
    + 
    \overbrace{
    \sum_{j=1}^{m} \bigg\| {\frac{\sum_{i=1}^{n} y_{i,j}  \pos_i}{\sum_{i=1}^{n}y_{i,j}} - \frac{\sum_{i=1}^{n} y_{i,j}  \goal_i}{\sum_{i=1}^{n}y_{i,j}}} \bigg\|_2
    }^{\text{Subteam-goals distance}}
\end{equation}
which jointly accounts for subteam adjacency, subteam balance, and subteam-goal distances.

\textbf{Intermediate-Level Graph Learning for Multi-robot Formation Adaptation}
To enable adaptive multi-robot formation control, 
we develop a graph learning approach at the intermediate level of STAF, 
which coordinates multiple robots to maintain a specific formation while adapting it based on the surrounding environment.
Given $\GGG$ that represents a team (or subteam) of robots along with the state $\state_i$ for each robot $i$, 
we develop a graph network $\phi$
to compute the embedding $\ff_i = \phi(\state_i, \GGG)$ of the team state with respect to the $i$-th robot,
which encodes the spatial relationships between the $i$-th robot with others in the team.
The network $\phi$ uses a linear layer to project the robot state $\state_i$
to the individual embedding $\zz_i$ of the $i$-th robot by $\zz_i = \WW^z \state_i$, 
where $\WW^z$ is the weight matrix of the linear layer.
Then, for the $i$-th robot,
 $\phi$ aggregates individual state embeddings of all other teammates through message passing to compute the team state embedding $\ff_i$ with respect to the $i$-th robot as $\ff_i = \WW^{f} \zz_i + \sum_{j \in \NNN(i)} \WW^{f} \left( \zz_j - \zz_i\right)$,
where $\WW^f$ is the weight matrix.
The team state embedding $\ff_i$ with respect to the $i$-th robot encodes
not only its own states (captured in the first term),
but also the relative spatial relationships with other teammates (captured in the second term), 
which facilitates the coordination of actions to maintain specific formations during multi-robot navigation.

Robot teams and subteams may encounter scenarios, such as narrow corridors, where rigidly maintaining their formations prevents successful navigation.
To enable formation adaptation, we implement a spring-damper model \cite{deng2024multi, gabellieri2021force} that dynamically adjusts the shape of the formation within the same type.
This spring-damper model includes two components:
(1) The spring component ensures that robot pairs maintain a balance between staying close enough to navigate narrow corridors and keeping a sufficient distance to prevent collisions, with the flexibility to adjust formation and enable adaptation.
This spring component is modeled as $|{d}_{i,j} - {p}_{i,j}|$, where ${d}_{i,j}$ denotes the expected distance in the original formation and ${p}_{i,j}$ represents the actual distance between the $i$-th and $j$-th robots, computed as $\|\pos_i-\pos_j\|_2$. 
(2) The damper component prevents oscillation and overshooting of each robot during navigation by smoothing the relative velocities between pairs of robots, which is defined as ${q}_{i,j} = \|\vel_i-\vel_j\|_2$. 
Combining these components, the spring-damper model for formation adaptation is mathematically defined as $R^\textit{adp} = \sum_{\mathbf{v}_i, \mathbf{v}_j \in \mathcal{V}} -\lambda |{d}_{i,j} - {p}_{i,j}| - (1-\lambda){q}_{i,j}$,
where $\lambda$ is a hyperparameter that balances the importance of the spring and damper components. 
$R^\textit{adp}$ is incorporated into the reward function, which is used to derive a loss function for  training STAF.






\textbf{Low-Level Individual Robot Control for Navigation}
At the low-level of STAF,
we introduce a navigation control network that outputs velocity commands as actions for each robot to reach its goal.
Given the state $\mathbf{s}_i$ for the $i$-th robot, we compute its state embedding $\ff_i$. 
We design the network $\psi$, which consists of two linear layers followed by the ReLU activation function, maps this embedding to an action as $\aaa_i=\psi(\ff_{i})$. 
The network $\psi$ is a part of the control policy $\pi_{\theta}(\aaa_i|\mathbf{s}_i)$, parameterized by $\theta$,
which is trained using the framework of reinforcement learning.
To enable each robot to move toward its target position and reach the navigational goal, 
we design a reward function based upon the distance between the current positions of the robot and its goal position.
To enable obstacle avoidance for safe navigation, we implement a reward function that imposes a penalty when a robot comes too close to nearby obstacles or other robots in the team.
When robots are divided into subteams, and once all subteams pass through the narrow corridor into an open area that is large enough for the full team,
the goal position of each individual robot 
is updated to align with the full team's goal, thereby recovering the subteams back into the full team with the original formation.

See Appendix for details on \textbf{STAF Training and Execution} with their time complexity analysis.

\section{Experiments}\label{sec:experiment}



\textbf{Experimental Setups}
We comprehensively evaluate our STAF approach across three setups: 
(1) a standard Gazebo simulation in ROS1, 
(2) a high-fidelity Unity-based 3D multi-robot simulator in ROS1, 
and (3) physical robot teams running ROS2.
Each setup involves different numbers and types of robots arranged in formations such as circle, wedge, and line.
In all scenarios, the environment includes narrow corridors, which require the full robot team to divide into subteams that adapt their formation to pass through. 
Afterward, the subteams regroup into the original full-team formation.
In simulation, robot poses and obstacles are obtained from Gazebo and Unity. 
In real-world experiments, robots use a SLAM approach \cite{zou2021comparative} for state estimation and mapping. 
See Appendix for details on approach implementation and training.
All video demonstrations are available on our project website.

We implement the complete STAF approach referred to as \textbf{STAF-full}. 
The full team divides into subteams to navigate through narrow environments, and after passing through, the subteams regroup into the full team to its original formation.
To analyze the performance of the subteams, we refer to the subteams as
 \textbf{STAF-sub\#}, e.g., STAF-sub1 and STAF-sub2.
For comparison with STAF, we further implement two previous methods for multi-robot coordinated navigation, including: 
(1) A Leader and Follower method (\textbf{L\&F}) \cite{reily2020leading} that one of the robots is designated as the ``leader robot'' that leads the movements of the other ``follower robots'' in the team while maintaining the formation.
(2) Decentralized GNN (\textbf{DGNN})  \cite{blumenkamp2022framework} that built upon a hierarchical learning framework to generate velocity controls for each individual robot for navigation, without considering team-level formations.

\begin{table*}[t]
\centering
\vspace{-4pt}
\tabcolsep=0.03cm
\caption{Quantitative comparison of STAF and Previous Methods from Gazebo simulations in ROS1.}
\label{tab:quant}
\vspace{4pt}
\tiny
\begin{tabular}{|l|c|c|c|c|c|c|c|c|c|c|c|c|c|c|c|}
\hline
Method  & \multicolumn{5}{c|}{Circle Formation} & \multicolumn{5}{c|}{Wedge Formation} & \multicolumn{5}{c|}{Line Formation}\\
\cline{2-16}
         & SR  ($\%$) & TT (sec)& $\sigma <0.5$ & $\sigma <0.1$ & $\sigma <0.01$ 
         & SR  ($\%$) & TT (sec)& $\sigma <0.5$ & $\sigma <0.1$ & $\sigma <0.01$
         & SR  ($\%$) & TT (sec)& $\sigma <0.5$ & $\sigma <0.1$ & $\sigma <0.01$
\\
\hline \hline
DGNN \cite{blumenkamp2022framework}  & \textbf{100.00} & {68.70} & 60.41 & 58.91    & 58.91 
         & \textbf{100.00} &82.70 & 47.85    & 42.33    & 41.92
         & \textbf{100.00} & 72.61 & 27.90 & 20.16 &  20.16
         \\
\hline
L$\&$F \cite{reily2020leading}  & 40.00 & \textbf{27.40} & 67.28 & 64.54 & 62.69 &                                                              70.00 & \textbf{26.50} & 69.70 & 62.11 & 59.47
                                & 60.00 & \textbf{30.10} & 63.76 & 55.89 & 55.89
\\

\hline
\hline
STAF-full     & \textbf{100.00} &102.10& \textbf{87.79} & \textbf{80.12} & \textbf{80.12} 
& \textbf{100.00} & 69.30 & \textbf{80.52} & \textbf{80.51} & \textbf{80.50}
& \textbf{100.00} & 111.50 & \textbf{91.45} & \textbf{80.06} &  \textbf{78.93}
 \\
\hline
\end{tabular}
\end{table*}

\begin{figure*}[t]
\centering
\vspace{-10pt}
\includegraphics[width=0.985\textwidth]{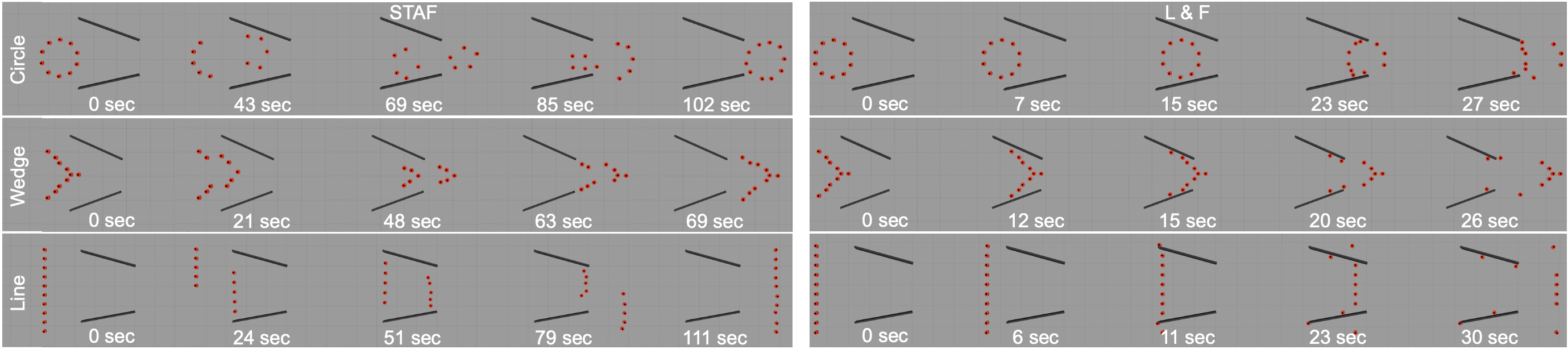}
\vspace{-6pt}
\caption{Qualitative results from Gazebo simulations on subteaming and formation adaptation.
}
\label{fig:qual}
\vspace{-6pt}
\end{figure*}

\begin{figure*}[!!!t]
\vspace{-9pt}
\centering
\subfigure[Circle formation]{\includegraphics[height=3cm]{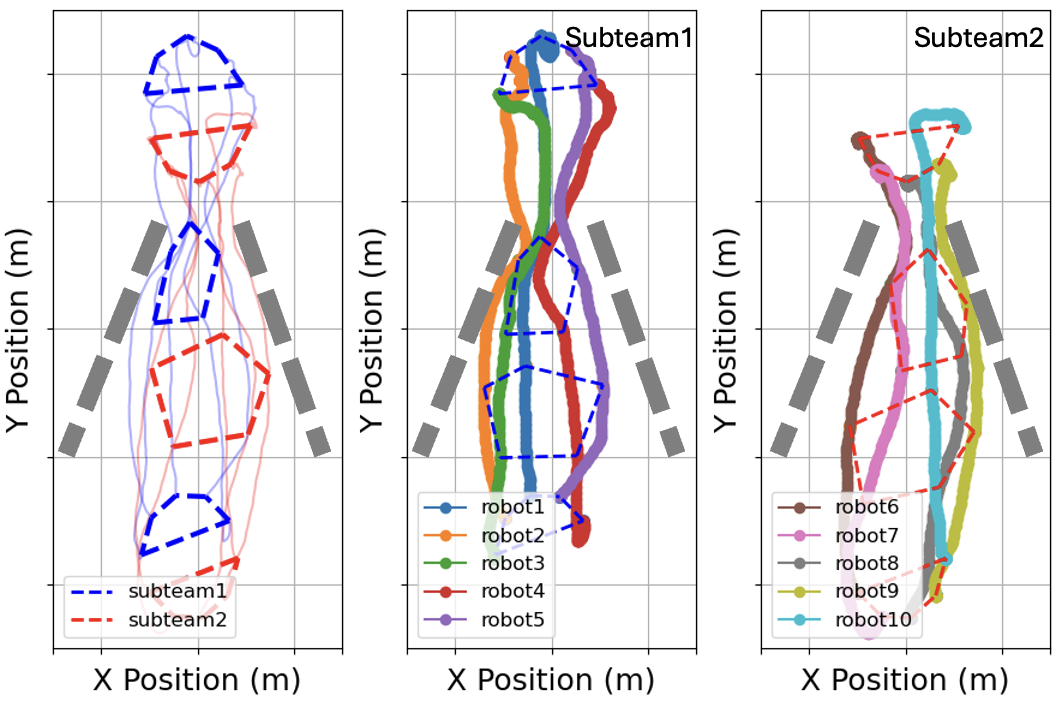}\label{fig:subcircle}}
\subfigure[Wedge formation]{\includegraphics[height=3cm]{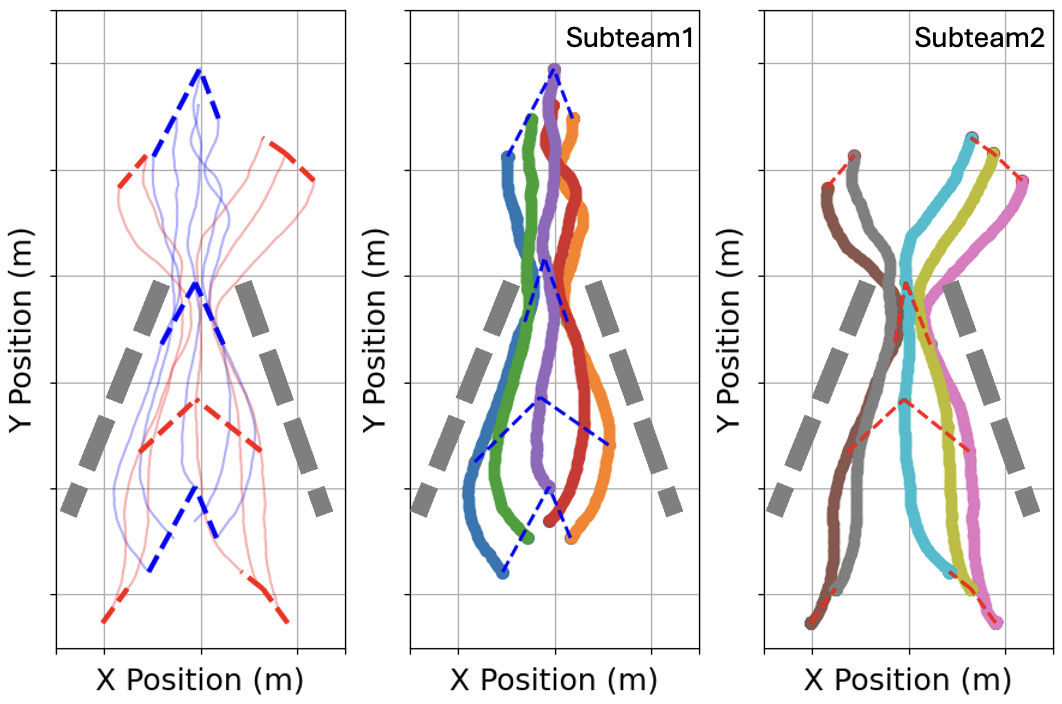}\label{fig:subwedge}}
\subfigure[Line formation]{\includegraphics[height=3cm]{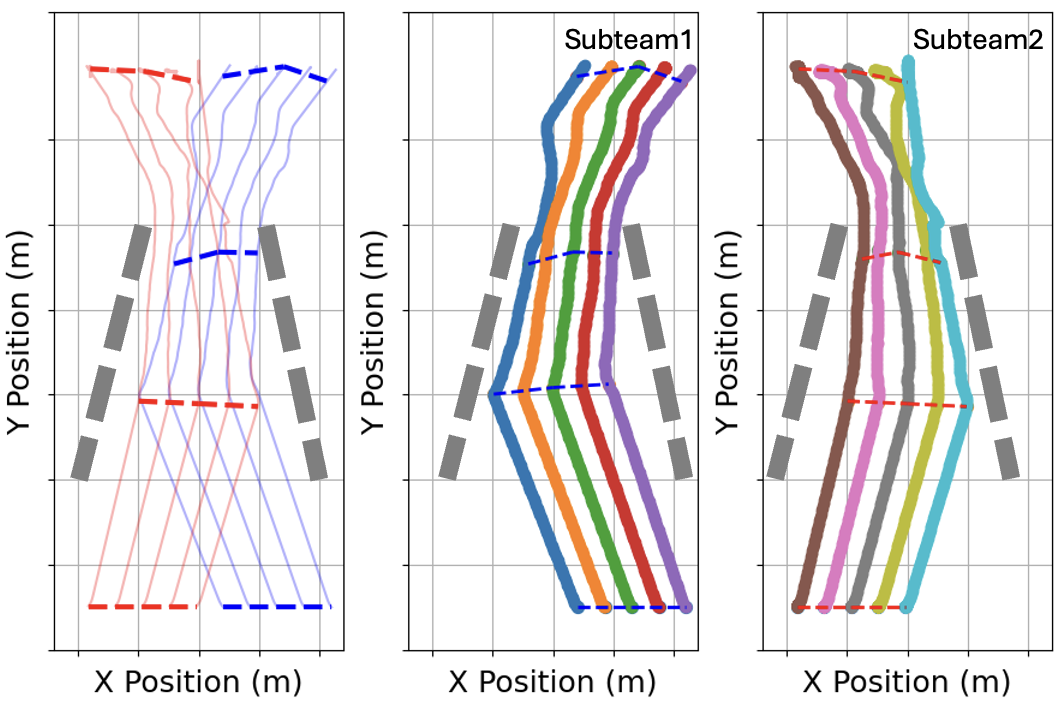}\label{fig:subline}}
\vspace{-6pt}
\caption{Movement trajectories of ten robots navigating a narrow corridor with different formations.
In Figure \ref{fig:subcircle} to \ref{fig:subline}, the first subfigure displays two subteams (red and blue) during team division, navigation with formation adaptation, and regrouping. 
The second and third subfigures show subteam trajectories, with each robot's path in a distinct color and gray dashed lines indicating obstacles.}
\vspace{-12pt}
\label{fig:trajectory}
\end{figure*}

To quantitatively evaluate and compare with other methods, we employ three metrics, including:
(1) Successful Rate (\textbf{SR}) is defined as the proportion of the robots within the full team that successfully reach goal positions without collisions. 
(2) Travel Time (\textbf{TT}) is defined as the total time used by the full team to reach the goal position. 
(3) Contextual Formation Integrity (\textbf{CFI}) is defined as the real-time adherence of the robots to their designated formation, given a shape threshold $\sigma$ that defines the strictness of the formation. The CFI $\in[0,1]$ evaluates how effectively a robot team utilizes the corridor gap and maintains formation, with smaller $\sigma$ indicating stricter formation requirements.
See Appendix for details on CFI and its calculation of different formations.

\textbf{Results in Multi-Robot Simulations}
The qualitative results in the Gazebo simulation are shown in Figure \ref{fig:qual}. 
{L$\&$F} gets stuck in the narrow corridor due to the lack of subteaming and formation adaptation. 
In contrast, our method autonomously divides the team, enabling each subteam to adapt formations and reach the goal; the first subteam starts moving, followed by the second, and they eventually merge into the full formation.
Notably, for wedge formations, team division prioritizes goal-distance objectives instead of maximizing connectivity, resulting in more compact subteams.

We visualize the trajectories of a team of 10 robots navigating in different formations, as shown in Figure \ref{fig:trajectory}. 
The visualization reveals subteaming behaviors (indicated by subteams in red and blue colors), including team division and regrouping. 
Additionally, formation adaptation of each subteam occurs when navigating through narrow corridors (indicated by the individual robot trajectories). 
These results show the effectiveness of STAF in enabling both subteaming and formation adaptation.

\begin{wraptable}{r}{0.43\textwidth} 
\centering
\tabcolsep=0.02cm
\tiny
\vspace{-20pt}
\caption{Quantitative results of two subteams from Gazebo simulations in ROS1.}
\label{tab:quant-subteam}
\vspace{6pt}
\begin{tabular}{|c|c|c|c|c|c|c|}
\hline
\multirow{2}{*}{Subteam} & \multirow{2}{*}{Formation} & \multicolumn{5}{c|}{Metrics} \\
\cline{3-7}
 &  & SR (\%) & TT (sec) & $\sigma <0.5$ & $\sigma <0.1$ & $\sigma <0.01$ \\
\hline \hline
\multirow{3}{*}{STAF-sub1} & Circle & 100.00 & 84.80 & 81.56 & 71.69 & 70.59 \\
 & Wedge  & 100.00 & 58.80 & 77.22 & 71.79 & 69.86 \\
 & Line   & 100.00 & 78.51 & 91.13 & 79.53 & 77.43 \\
\hline
\multirow{3}{*}{STAF-sub2} & Circle & 100.00 & 59.50 & 87.72 & 82.67 & 80.11 \\
 & Wedge  & 100.00 & 46.80 & 80.99 & 80.15 & 75.79 \\
 & Line   & 100.00 & 90.06 & 91.78 & 81.74 & 80.43 \\
\hline
\end{tabular}
\vspace{-10pt}
\end{wraptable}
The quantitative results are shown in Table \ref{tab:quant}. DGNN performs the worst, particularly in the CFI metrics, as it lacks formation control. 
L$\&$F uses formation control and performs better but has only a 40$\%$ success rate, as it considers rigid formation control. 
Our method outperforms both by addressing these limitations, which achieves a 100$\%$ success rate.
Although STAF yields slightly longer travel times, this is expected due to its more complex navigation strategy.
Subteam performance in Table \ref{tab:quant-subteam} shows a 100$\%$ success rate across all formations. 
STAF maintains formation integrity above 87$\%$, 80$\%$, and 91$\%$ under the threshold $\sigma < 0.5$ for circle, wedge, and line formations, respectively. These results highlight the effectiveness of STAF in enabling coordinated navigation through subteaming and adaptive formation control.

\begin{figure*}[ht]
\vspace{-6pt}
\centering
\includegraphics[width=0.985\textwidth]{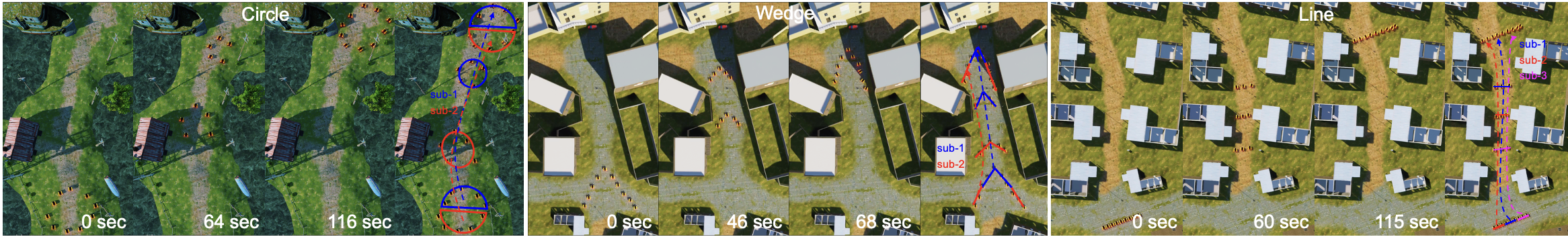}
\vspace{-6pt}
\caption{Qualitative results from Unity3D simulations in ROS1 using varying numbers of differential-drive Warthog robots in three formations while navigating a long, unstructured field environment.}
\label{fig:3d}
\vspace{-6pt}
\end{figure*}

\begin{figure*}[ht]
\vspace{-6pt}
\centering
\includegraphics[width=0.985\textwidth]{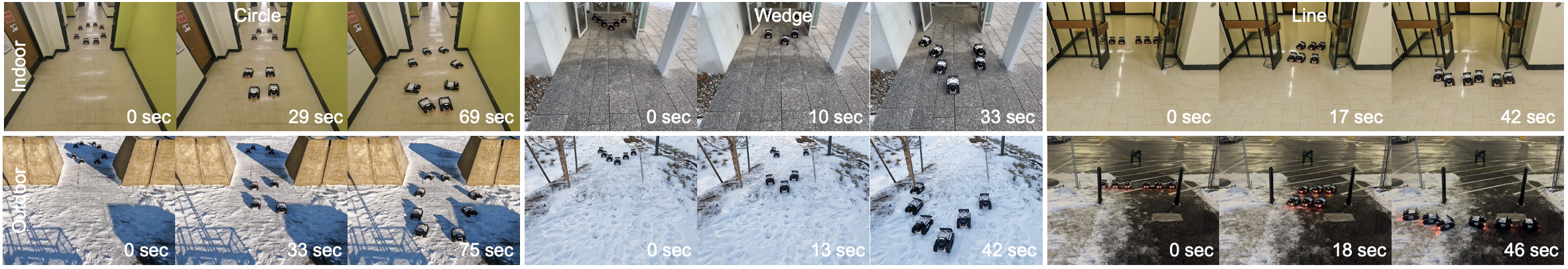}
\vspace{-6pt}
\caption{Qualitative results from real-world experiments in both indoor narrow spaces and outdoor uneven terrain, using varying numbers of Limo robots running ROS2 and communicating via Wi-Fi.}
\label{fig:real}
\vspace{-6pt}
\end{figure*}
Beyond the Gazebo simulation, we further use a high-fidelity Unity3D simulator in ROS1, which simulates outdoor field environments with narrow pathways and bridges.
Instead of using holonomic robots as shown in the Gazebo simulation, we use differential-drive Warthog robots and convert the linear velocity in the action $\mathbf{a}_i$ into wheel velocities to follow the same trajectory. 
This setting introduces new challenges, which require the robot team to navigate long curved paths that demand continuous formation adjustments.
As illustrated in Figures \ref{fig:3d}, our STAF approach successfully addresses these challenges by dividing a full team into subteams, adapting actions of differential-drive subteams to navigate, and regroup after subteam traversal. For line formation with 9 robots, STAF can  divide into three subteams to navigate a corridor too narrow for groups larger than 3. 


\textbf{Case Study on Physical Robot Teams}
We validate STAF on real-world case studies using differential-drive Limo robots with caterpillar tracks, each equipped with an onboard Intel NCU i7 and running ROS2 with Wi-Fi-based team communication.
The real-world experiments are conducted both indoors and outdoors, 
as shown in Figure \ref{fig:real}. 
Our method enables teams of 6 to 8 robots to divide into subteams and adapt formations to smoothly navigate narrow indoor spaces. 
In outdoor experiments on unstructured terrain, the results demonstrate the strong adaptability of our approach to unknown environments. Subteaming and formation adaptation are effectively performed even on snowy and uneven terrain, where wheel slippage introduces significant action uncertainty. 
Additional Unity3D and real-world qualitative results with more timesteps are provided in the Appendix.

\section{Discussion}
\begin{figure*}[!!!t]
\centering
\vspace{-12pt}
\subfigure[ST-B]{\includegraphics[width=0.15\linewidth]{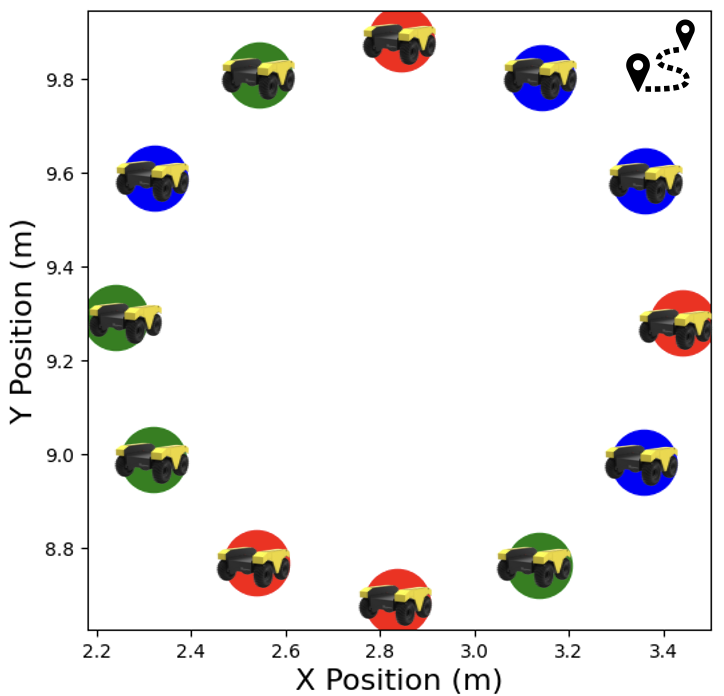}\label{fig:disbalance}}
\hfill
\subfigure[ST-A]{\includegraphics[width=0.15\linewidth]{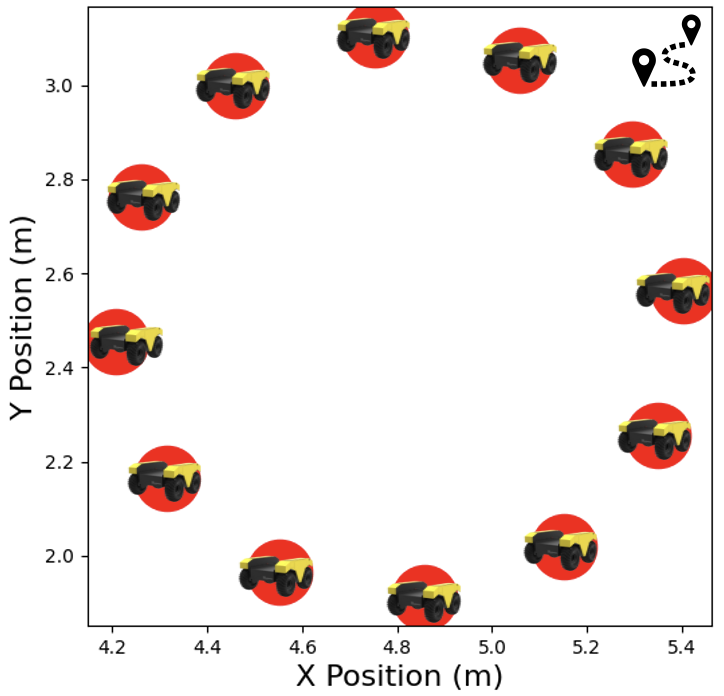}\label{fig:dis3connect}}
\hfill
\subfigure[ST-G]{\includegraphics[width=0.15\linewidth]{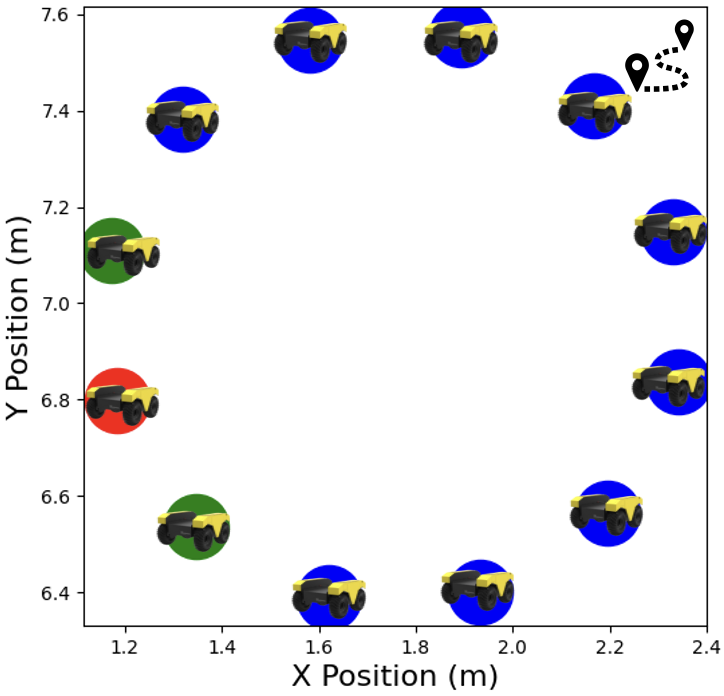}\label{fig:dis4goal}}
\subfigure[w/o ST-B]{\includegraphics[width=0.15\linewidth]{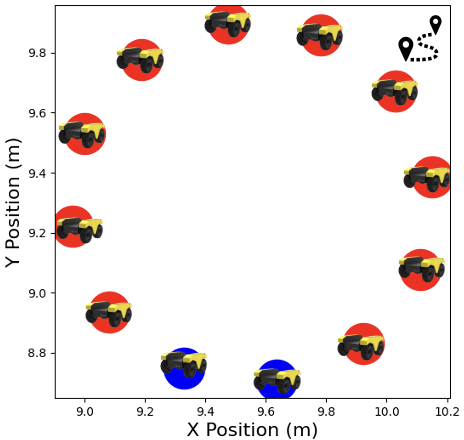}\label{fig:disdico}}
\hfill
\subfigure[w/o ST-A]{\includegraphics[width=0.15\linewidth]{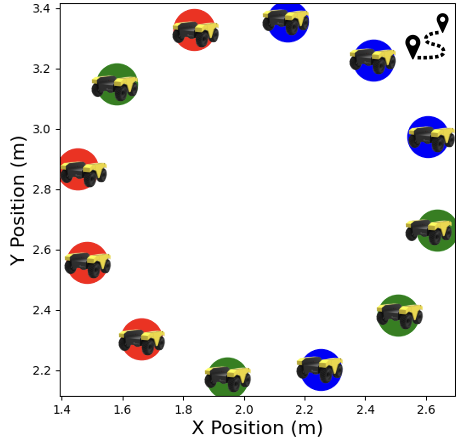}\label{fig:disdiba}}
\hfill
\subfigure[w/o ST-G]{\includegraphics[width=0.15\linewidth]{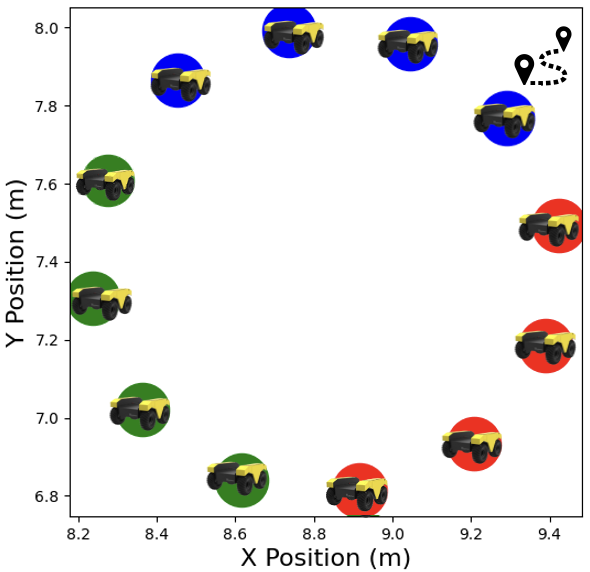}\label{fig:disbaco}}
\vspace{-6pt}
\caption{Ablation study that analyzes the impact of subteam division components: subteam balance (ST-B), subteam adjacency (ST-A), and subteam-goals distance (ST-G). }
\label{fig:dis_graph_loss}
\end{figure*}

\begin{figure*}[!!!t]
\centering
\vspace{-12pt}
\subfigure[4 robots]{\includegraphics[height=3.7cm]{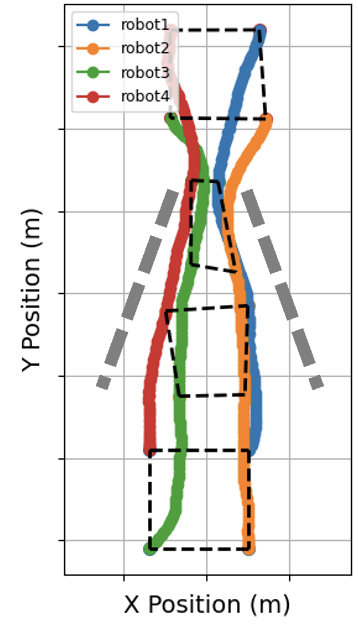}\label{fig:dis4}}
\subfigure[6 robots]{\includegraphics[height=3.7cm]{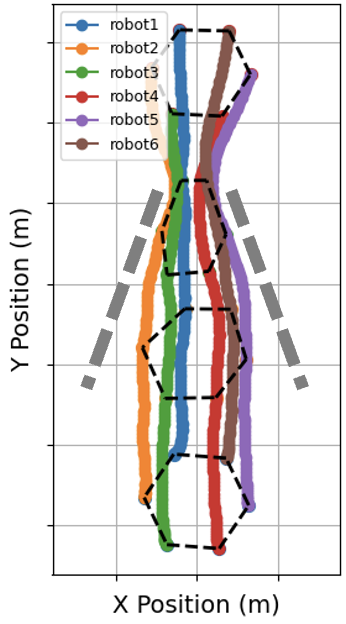}\label{fig:dis6}}
\subfigure[7 robots]{\includegraphics[height=3.7cm]{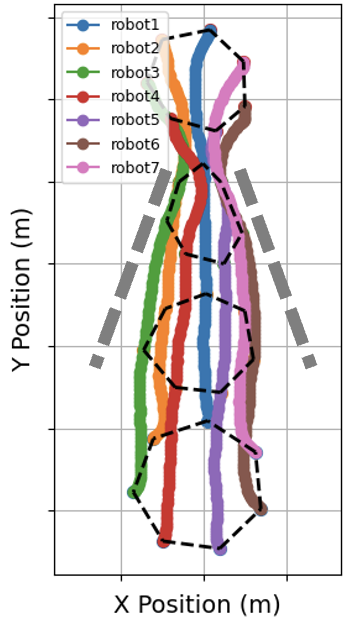}\label{fig:dis7}}
\subfigure[8 robots]{\includegraphics[height=3.7cm]{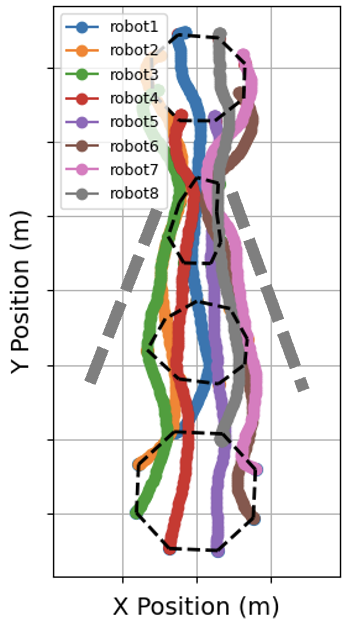}\label{fig:dis8}}
\subfigure[CFI across team sizes]{\includegraphics[height=3.7cm]{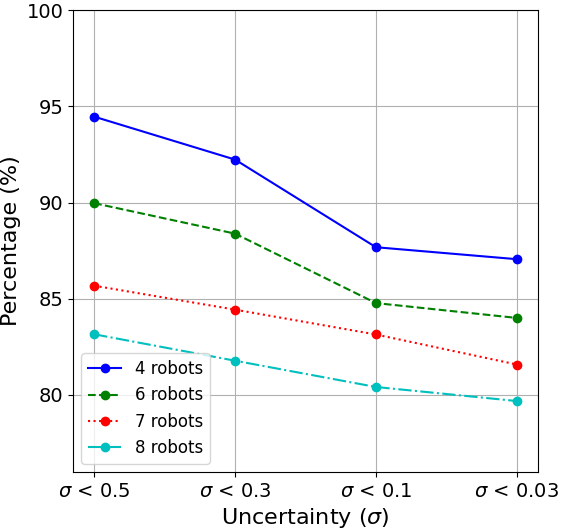}\label{fig:discompare}}
\vspace{-6pt}
\caption{Quantitative results indicate STAF's generalizability to different team sizes. Figures (a)-(d) show the trajectories of 4 to 8 robots in circle formations to navigate a narrow corridor. 
Figure (e) presents the variation in CFI values across different team sizes and $\sigma$ values.}
\label{fig:dis}
\vspace{-12pt}
\end{figure*}

\textbf{Ablation Study on Subteam Division} 
We conduct an ablation study to evaluate the role of each component in the objective function defined in Eq. (\ref{eq:loss}) for team division. 
Figure \ref{fig:disbalance} shows that optimizing only the balance term evenly splits 12 robots into 3 subteams.
Figure \ref{fig:dis3connect} shows that only maximizing adjacency leads to all robots being assigned to the same subteam.
Figure \ref{fig:dis4goal} shows that only minimizing the goal-distance aligns subteams toward their goals (in the upper right). 
In addition, we remove each component individually to assess its impact.
Figure \ref{fig:disdico} shows unbalanced team division without the balance term.
Figure \ref{fig:disdiba} results in uncompact subteams without the adjacency term. 
Figure \ref{fig:disbaco} shows subteams misaligned with goals, which leads to inefficient navigation.
These results further indicate the effectiveness and importance of enforcing subteam balance, maximizing adjacency, and minimizing subteam-goals distance for robot team division.


\textbf{Generalizability to Different Team Sizes} 
We evaluate the generalizability of STAF to different team sizes by varying the number of robots. 
Figures \ref{fig:dis4}-\ref{fig:dis8} present the qualitative results on formation adaptation for teams of 4, 6, 7, and 8 robots in circle formation, which validate STAF's generalizability across team sizes. 
Figure \ref{fig:discompare} presents the quantitative results using the CFI metric, which shows 87$\%$ formation integrity for 4 robots under $\sigma < 0.03$, and at least 80$\%$ for 8 robots.

\begin{wrapfigure}{R}{0.50\textwidth}
\centering
\vspace{-19pt}
\subfigure[2 subteams]{\includegraphics[width=0.3\linewidth]{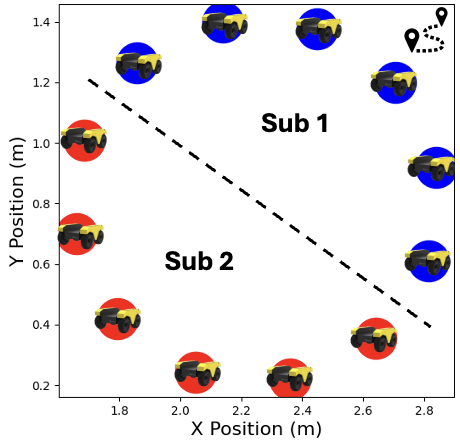}\label{fig:dis2sub}}
\hfill
\subfigure[3 subteams]{\includegraphics[width=0.3\linewidth]{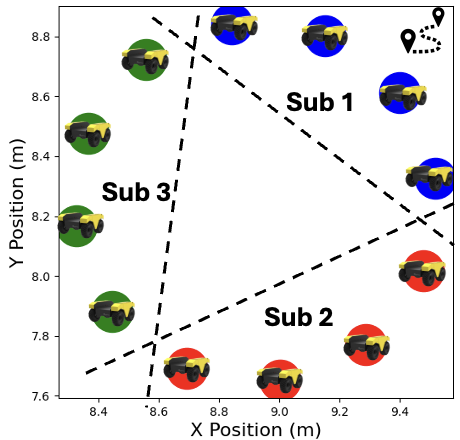}\label{fig:dis3sub}}
\hfill
\subfigure[4 subteams]{\includegraphics[width=0.3\linewidth]{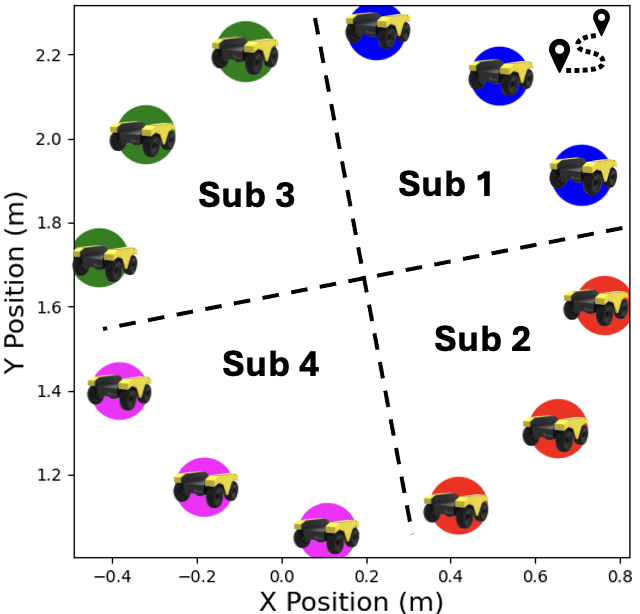}\label{fig:dis4sub}}
\vspace{-8pt}
\caption{Qualitative results indicate STAF's generalizability to different numbers of subteams.}
\label{fig:dis_graph}
\vspace{-12pt}
\end{wrapfigure}

\textbf{Generalizability to Different Numbers of Subteams} 
We evaluate STAF’s generalizability in dividing the team into varying numbers of subteams. As shown in Figure \ref{fig:dis_graph}, STAF effectively handles divisions into 2, 3, and 4 subteams. Figure \ref{fig:3d} contains a scenario where a nine-robot line formation splits into three subteams to navigate a corridor too narrow for groups larger than four.


See Appendix for STAF's \textbf{Robustness to Noise} and \textbf{Applicability to Different Robot Platforms}. 

\section{Conclusion}\label{sec:conclusion}

In this paper, we propose STAF for coordinated multi-robot navigation in complex scenarios.
STAF is built upon a unified hierarchical learning framework, 
including a high-level deep graph cut for dynamic team division, 
an intermediate-level graph learning for team coordination with adaptive formation control, 
and a low-level RL policy for individual robot control.
Results from comprehensive experiments show that STAF enables new multi-robot capabilities for subteaming and formation adaptation, and significantly outperforms existing methods on coordinated multi-robot navigation.


\clearpage
\section{Limitations}\label{sec:limitation}


Our approach presents several limitations that suggest directions for future research.  
First, 
although STAF's intermediate and low levels are executed in a decentralized fashion,
STAF's high level for team division is executed in a centralized fashion.
One direction for future research is to decentralize the high-level team division, 
such as by replacing the current global graph cut optimization with a distributed consensus algorithm (e.g., gossip \cite{boyd2006gossip} or max-consensus \cite{olfati2007consensus}). These decentralized methods would enable each robot to determine its subteam 
based upon the information shared by its teammates through broadcasting, 
and iteratively reach a consensus and converge to a stable subteam assignment through negotiation.
Second, the alternating training algorithm we use, which iteratively trains the high-level and joint intermediate-low levels, is considered a limitation,
as it may lead to suboptimal integration of these levels and difficulties with training error propagation. 
In the future, we plan to integrate the high-level graph cut together with the joint intermediate-low level training into an end-to-end training algorithm,
where the training error from the low level will be propagated not only to the intermediate level but also to the high level,
which enables updates to the network parameters across all three levels.
To achieve this, we will adopt a centralized training with decentralized execution strategy, where all levels of the hierarchy can leverage global information during training, while ensuring decentralized execution during deployment.
The third limitation is that the number of subteams, as a hyperparameter, is decided manually. A future direction is to dynamically and adaptively determine this hyperparameter by selecting the minimum number of subteams such that the smallest formation of each subteam can successfully navigate through the narrowest corridor in the environment. 
The width of a corridor can be identified either by analyzing the environment map (using a prior map or built by a SLAM method) or through real-time robotic sensing.
\acknowledgments{This work was partially supported by the NSF CAREER Award IIS-2308492, DEVCOM ARL TBAM CRA W911NF2520024, and the DARPA Young Faculty Award (YFA) D21AP10114-00.}


\bibliography{references}  
\end{document}